\documentclass[letterpaper, 10 pt, conference]{ieeeconf}  

\IEEEoverridecommandlockouts                              

\usepackage{amsmath} 
\usepackage{amssymb}  

\usepackage{graphicx}
\usepackage{capt-of} 
\usepackage{overpic} 
\usepackage{minibox} 
\usepackage{textcomp} 
\usepackage{tabularx}
\usepackage{multirow}
\usepackage{booktabs}
\usepackage{arydshln}
\usepackage{siunitx}
\usepackage{pifont}
\usepackage{stfloats}
\usepackage{makecell}
\usepackage{wrapfig}
\usepackage{url}
\usepackage{graphicx}
\usepackage{array}
\usepackage{multicol}
\usepackage{multirow}
\usepackage{geometry}
\usepackage{silence}
\usepackage{subcaption}
\usepackage{csquotes} 
\usepackage{fancyhdr}
\usepackage{nicematrix}  
\usepackage{algpseudocode,algorithm,algorithmicx}
\usepackage[normalem]{ulem}
\usepackage{xspace}
\usepackage{cite}
\usepackage{afterpage}

\algrenewcommand\algorithmicrequire{\textbf{Precondition:}}
\algrenewcommand\algorithmicensure{\textbf{Postcondition:}}

\makeatletter
\DeclareRobustCommand\onedot{\futurelet\@let@token\@onedot}
\def\@onedot{\ifx\@let@token.\else.\null\fi\xspace}

\makeatother

\DeclareSIUnit{\million}{\text{Mio.}}

\makeatletter
\let\NAT@parse\undefined
\makeatother
\usepackage[pagebackref,breaklinks,colorlinks,allcolors=blue]{hyperref}

\title{\LARGE \bf
AutoVDC: Automated Vision Data Cleaning\\
Using Vision-Language Models
}
\author{Santosh Vasa$^{*1}$, Aditi Ramadwar$^{*1}$, Jnana Rama Krishna Darabattula$^{1}$,\\
Md Zafar Anwar$^{1}$, Stanislaw Antol$^{1}$, Andrei Vatavu$^{1}$, Thomas Monninger$^{1,2}$, Sihao Ding$^{1}$
\thanks{$^{*}$Authors denoted with * contributed equally to this paper.}%
\thanks{$^{1}$Mercedes-Benz Research \& Development North America, San Jose, CA, USA}%
\thanks{$^{2}$University of Stuttgart, Institute for Artificial Intelligence, Stuttgart, Germany}%
}

\begin{document}
\bstctlcite{IEEEexample:BSTcontrol}
\maketitle
\thispagestyle{empty}
\pagestyle{empty}
\fancyhf{}

\begin{abstract}
Training robust autonomous driving systems requires extensive datasets with precise annotations, yet manual data curation is expensive and human labels are often imperfect. 
In this paper, we conduct a systematic study of Vision-Language Model (VLM) maturity for automated data curation.
To derive our findings, we utilize AutoVDC (Automated Vision Data Cleaning), a model-agnostic, two-stage framework that leverages the generalization gap of neural networks to audit large-scale datasets. 
In the pipeline, a task model first proposes suspicious candidates, which a VLM then acts as a judge to validate errors. 
Using this framework, we evaluate the readiness of VLMs for annotation cleaning on KITTI and nuImages. 
Our study reveals that zero-shot VLMs are effective on semantic errors but show a clear spatial reasoning gap for localization noise, making Fine-Tuning with Chain-of-Thought (FT-CoT) essential. 
We further demonstrate that across variants with intentionally injected erroneous annotations, this VLM-based auditing approach remains robust even when guided by a weak proposer and successfully uncovers 39 previously unknown errors in the original KITTI ground truth.

\end{abstract}


\section{Introduction}\label{sec:intro}

In machine learning, high-quality and efficient data annotation is crucial for developing effective models. 
Traditional manual annotation presents significant challenges, including a large workload, inconsistent quality, and a high cost. 
These issues limit the scalability and quality of annotations, adversely affecting model performance and evaluation.
While auto-labeling algorithms have improved data collection efficiency by automating parts of the annotation process \cite{Chen_2022}, they often introduce systemic biases that are  problematic in high-stakes domains such as autonomous driving, where continuous refinement and validation are required.

These challenges are amplified by the rapid growth of dataset scale.
Public dataset sizes in autonomous driving have expanded substantially, as seen in the Waymo Open Dataset \cite{Sun_2020_CVPR} and Argoverse \cite{wilson2023argoverse2generationdatasets}. Industry and proprietary datasets are growing at an even faster rate, driven by the need for more comprehensive and diverse data to train robust models. This rapid growth makes label-by-label manual inspection increasingly costly and difficult to sustain, motivating the development of automated approaches that can scale.
\begin{figure}[t]
\centering
\vspace{-0.2cm}
\includegraphics[width=0.5\textwidth, trim=20 180 160 0, clip=true]{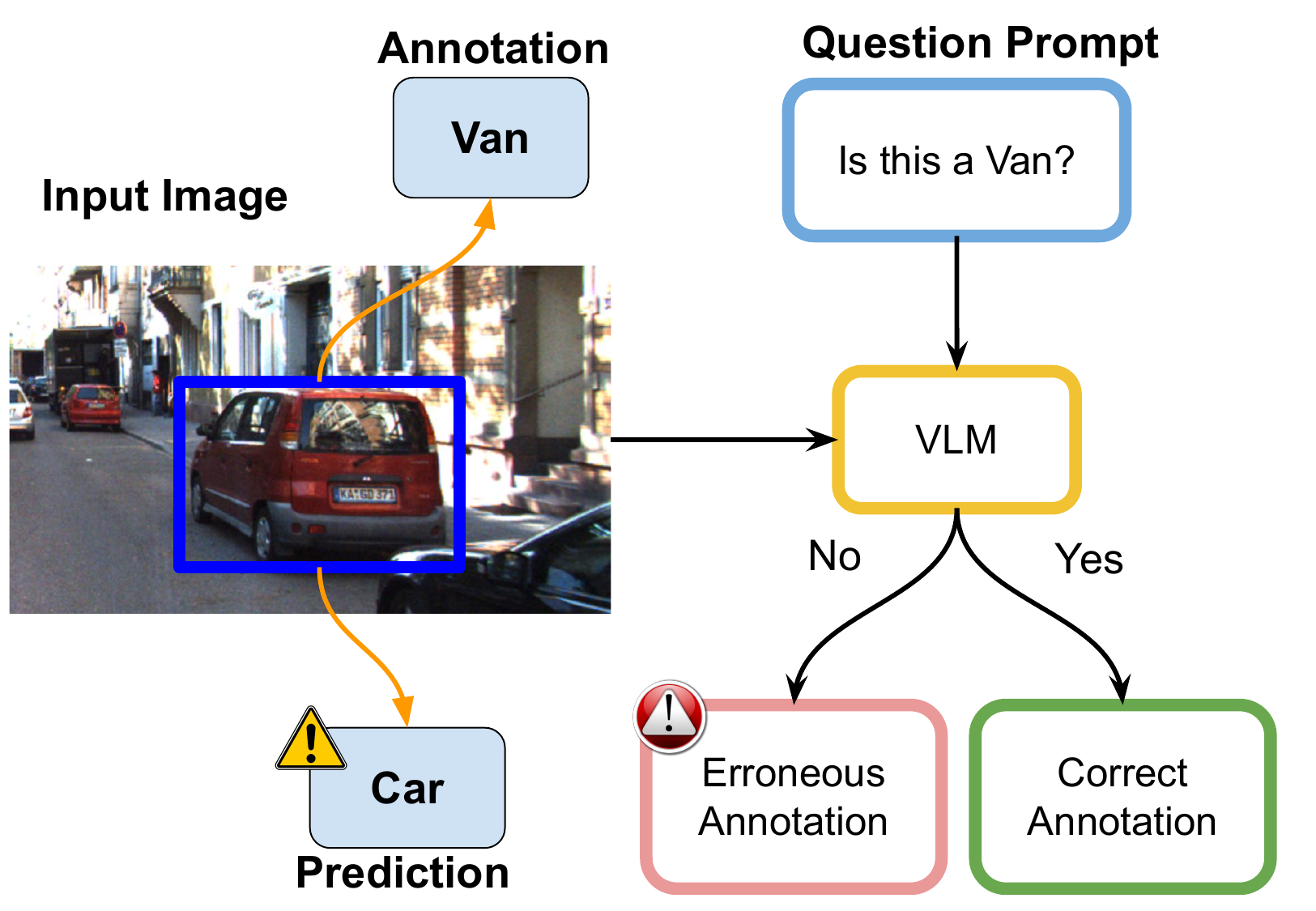}
\caption{
The two-stage auditing process where the system identifies a label-prediction conflict (Error Proposal) and utilizes a VLM to validate the annotation’s correctness (Error Validation).
}
\vspace{-0.2cm}
\label{fig:teaser}
\end{figure}

In response to the above limitations, we conduct a systematic readiness study to evaluate the maturity of Vision-Language Models (VLMs) for automated data curation.
To derive our findings, we utilize
\textbf{AutoVDC} (\textbf{Auto}mated \textbf{V}ision \textbf{D}ata \textbf{C}leaning),
which we define as a two-stage label cleaning method, Fig. \ref{fig:teaser}, that integrates VLMs to automate the detection of annotation errors. 
The first stage acts as a triage stage that flags potential error candidates from the entire dataset. 
A VLM in the 
second
stage assesses these proposals to determine whether or not the annotations are correct, which eliminates or substantially reduces the need for human intervention and effort. 

By using
AutoVDC as a test framework, we address a broader question: How mature are 
VLMs for high-precision perception auditing? While VLMs demonstrate strong open-ended understanding\cite{vqa2015},  perception datasets demand fine-grained spatial inspection in addition to semantic correctness. This motivates our study, focused on both the strengths and limitations of current VLMs when used as automated dataset auditors.

To evaluate the effectiveness of VLM-based curation, this paper addresses the following research questions (RQ):


\noindent \textbf{RQ1. VLM Maturity and the Spatial Reasoning Gap.} 
To what extent do current VLMs exhibit the architectural maturity required to resolve fine-grained localization noise compared to semantic classification errors, and is domain alignment through fine-tuning essential to bridge the remaining spatial reasoning gap?

\noindent \textbf{RQ2. Exploiting the Generalization Gap.} 
Does the tendency of task models to learn general semantic features before memorizing noise provide a reliable signal for automated error discovery, even when the task model (model specific to a task) is trained on noisy labels?

\noindent \textbf{RQ3. The Resilience of Triage Logic.} 
Can the VLM act as a robust gatekeeper to maintain high system precision even when utilizing a weak task model with high false-positive rates in the triage stage?


Our contributions are directed towards answering these questions and showcasing the analysis of our findings.






\begin{itemize}

    \item A VLM readiness study: 
    A systematic evaluation on KITTI \cite{Geiger2012CVPR} and nuImages \cite{nuscenes} showing that while zero-shot VLMs work well for semantic errors, they exhibit a spatial reasoning gap; we show that FT-CoT is essential to resolve fine-grained localization errors.
    
    \item The AutoVDC framework: 
    A modular, two-stage methodology that combines task-model triage with VLM-based validation for scalable, high fidelity annotation cleaning.
    
    \item Robustness and Real-World Impact:
    We demonstrate that VLM-based auditing is resilient
    to weak proposers, and 
    successfully uncovers 39 previously unknown errors in the original KITTI ground truth, improving evaluation fidelity.
    
\end{itemize}

\section{Related Work}\label{sec:related_work}
To address the challenge of producing accurately labeled datasets, numerous data-driven methodologies have been introduced.
For instance, anomaly detection is employed in label cleaning to identify data points that deviate significantly from the norm, thereby flagging samples that are likely to contain labeling inaccuracies \cite{bishop2006pattern, ramaswamy2000efficient}.
Moreover, several advanced methods integrate multiple strategies to improve the identification of erroneous annotations.
Standard error detection utilizes model ensembling, entropy-based measures, and active label cleaning \cite{Bernhardt_2022,lakshminarayanan2017simple,lewis1994sequential, settles2009active} to identify prediction-annotation divergences. 
Although these methods effectively highlight potential label inaccuracies, they still require human review to confirm the flagged samples, making the process labor-intensive.
Studies like \cite{heyn2023automotiveperceptionsoftwaredevelopment} and \cite{liu2024surveyautonomousdrivingdatasets} have examined the challenges, costs, and efforts associated with these existing approaches, which we aim to obviate.

In recent years, generative models like Large Language Models (LLMs) \cite{zheng2023judgingllmasajudgemtbenchchatbot} have increasingly replaced components of machine learning development in both academia and industry. LLMs are now used to evaluate models \cite{li2024llmsasjudgescomprehensivesurveyllmbased} and generate synthetic data and annotations \cite{guo2024generativeaisyntheticdata}, significantly reducing the need for human labor.
More recently, Large Vision Models (LVMs), such as Vision Foundation Models (VFMs) \cite{radford2021learningtransferablevisualmodels, xiao2023florence2advancingunifiedrepresentation, ravi2024sam2segmentimages} and VLMs \cite{agrawal2024pixtral12b, liu2023visualinstructiontuning, abdin2024phi3technicalreporthighly}, have markedly improved the ability of machines to perceive and understand visual information.
These models have demonstrated impressive generalist capabilities \cite{agrawal2024pixtral12b}, enabling them to perform a wide range of tasks across various domains in both vision and multi-modal contexts \cite{xiao2023florence2advancingunifiedrepresentation, liu2023visualinstructiontuning}.
This versatility is attributed to their training on large, extensive and diverse vision and language datasets \cite{schuhmann2022laion5bopenlargescaledataset} that are significantly larger than task-specific model datasets \cite{nuscenes}.
This allows VLMs to understand and process relationships across modalities, thereby enhancing their performance in tasks that require both textual and visual information, thus ideal for VQA applications.
These models can also reason image markups, such as boxes or arrows overlaid onto the image, that can be referenced in the text input \cite{cai2024vipllavamakinglargemultimodal}.

Recent work is starting to leverage these abilities for active learning and label cleaning.
\cite{Kim_2024_CVPR} utilized VLMs to generate pseudo-labels in an incremental learning approach to train object detection models.
\cite{diab2025vlmsforsemisupervisedlearning} employed VLMs to validate pseudo-labels specialized for geo-locating voltage cabins in street view imagery in a semi-supervised setting.
ClipGrader \cite{lu2025clipgraderleveragingvisionlanguagemodels} fine-tuned CLIP to identify labeling errors in object detection datasets by determining similarity between textual and visual features. However, ClipGrader \cite{lu2025clipgraderleveragingvisionlanguagemodels} is specialized for object detection and the use of the CLIP model, requiring substantial modifications to integrate large VLMs. In contrast, AutoVDC is modular, capable of integrating various VLMs, and can be extended to vision tasks other than object detection with minimal modifications. Additionally, with triage stage, AutoVDC framework drastically reduces the amount of annotations processed by a VLM, thereby aiding computational efficiency. 

Nevertheless, utilization of VLMs in the ML development cycle, particularly for dataset cleaning, is an emerging area of study. 
In this work, we present a modular system that can leverage the wide range of VLMs to automatically identify erroneous annotations.

\section{Approach}\label{sec:approach}


\subsection{Problem Statement}\label{sec:problem_statement}

The goal of automated data cleaning is to identify a subset of erroneous annotations $D^{\prime}$ within an imperfect dataset of $N$ data points, ${D = \{(a_i, x_i)\}_{i=1}^N}$, where $x_i$ represents an input image and $a_i$ is the corresponding singular annotation (e.g., a bounding box or class label). We define a classification function
${f ({a_i}, {x_i}) \to \quad {y_i} \in \{\mathrm{yes}, \mathrm{no}\}}$
that maps each sample to its true correctness.

\begin{figure*}[ht]
\centering
\vspace{-0.1cm} 
\includegraphics[width=0.97\textwidth, trim=3 365 2 125, clip=true]{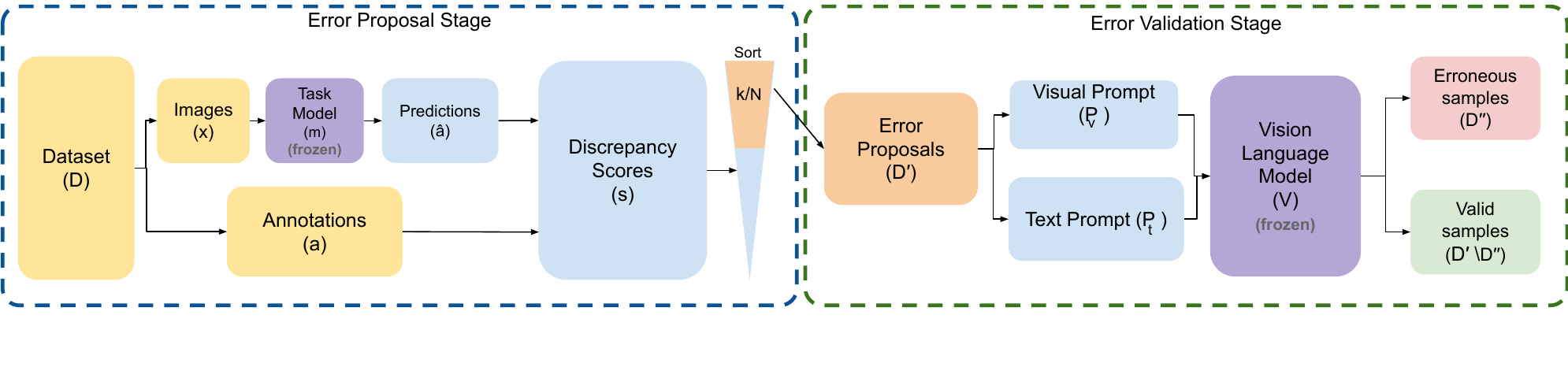}
\vspace{-0.6cm}
\caption{Illustrates overall system architecture.
\textit{Left}: Error Proposal Stage scores samples based on model predictions and annotations.
\textit{Right}: Error Validation Stage employs a VLM to examine proposals and detect annotation errors. 
}
\vspace{-0.3cm}
\label{fig:overall_architecture}
\end{figure*}

Our implementation, 
visualized in Fig.~\ref{fig:overall_architecture},
approximates $f$ by first exploiting the generalization gap of task models to propose discrepancies and then utilizing the spatial reasoning of VLMs to audit these proposals. This dual-intelligence approach allows the system to distinguish between inaccurate task model predictions and genuine annotation decay, providing a scalable alternative to exhaustive manual review.

\subsection{Error Proposal (EP) Stage}\label{sec:discrepancy_scoring}

The EP stage serves as a high-recall filter designed for computational efficiency. Its primary goal is to reduce the volume of data passed to the VLM by identifying a candidate subset that likely contains errors.

We utilize a perception task model $m_{\theta}$ to generate predictions $\hat{a}_i$. Even if $m_{\theta}$ is trained on noisy data, it typically learns consistent semantic features before memorizing random noise. We exploit this generalization gap by calculating a discrepancy score $s_{i}$ = {$\delta({\hat{a}_i}, {a_i})$}, where $\delta$ is a task-specific function representing the degree of discrepancy (e.g., IoU, cosine similarity).

We then select the top-k samples or those exceeding a score threshold to form a set of error proposals $D\prime$. This process addresses RQ2 by concentrating the system's attention on suspicious samples, reducing the VLM's total workload significantly.

\subsection{Error Validation (EV) Stage}\label{sec:vlm_error_classification}

The EV stage acts as the final judge, using the reasoning power of a VLM, $V$, to determine the true state of each proposal $(a_i, \hat{a}_i, x_i) \in D'$ by producing $\hat{y}_i$. The VLM is used as an expert auditor. 
We provide the VLM with two inputs, as seen in Fig. \ref{fig:teaser}:

1) Visual Prompt $P^v$: A crop of $x_i$, centered on the discrepancy, overlaid with the bounding box $a_i$ or $\hat{a}_{i}$ to focus the model's spatial attention.

2) Text Prompt $P^t$: A specific query requiring the VLM to reason whether the annotation $a_i$ accurately describes the visual content of $x_i$.




The VLM generates a response,  ${\hat{y}_i = V(P^v_i, P^t_i)}$, that determines the correctness of the annotation.
From $D^{\prime}$, a new subset ${D^{\prime\prime}}$ is created consisting of validated erroneous annotations.
This verified set can be used to eliminate (Section~\ref{sec:experiments}) or correct the flagged labels.


\subsection{Application to 2D Object Detection} \label{sec:app_to_obj_det}


To demonstrate the framework’s effectiveness, we instantiate AutoVDC for the task of 2D object detection.


\textbf{EP Stage}: 
Discrepancies are identified by matching predictions to ground-truth annotations based on an Intersection over Union (IoU) threshold of 0.65. Adjusting this threshold enables finer or coarser selection of samples for processing in the EV stage. Objects that fail to meet the matching criterion are marked as discrepancies with a score of $s_i = 1$. 
These cases fall into three categories:


\begin{enumerate}
\item \textbf{Class mismatch}: The model and annotation agree on the location but the object class categories differ.
\item \textbf{Missing prediction}: An annotation exists without a matching prediction.
A potential false negative in task evaluation.
\item \textbf{Missing annotation}: The model predicts a high-confidence object where no annotation exists. A potential false positive in task evaluation.
\end{enumerate}
For simplicity, we assign $s_i = 1$ to all such cases, treating them as equally discrepant, and include them in the set $D^\prime$.

\textbf{EV Stage}: Our visual prompt $P^v$ is created by overlaying the bounding box of interest
on the image $x_i$ and then cropping the image around the box of interest with fixed padding.
Our textual prompt $P^t$ uses discrepancy information to substitute in for placeholders to produce the final prompts, as shown in the right side of Fig ~\ref{fig:teaser}.

The VLM then acts as an expert auditor, reasoning whether the box accurately fits the visual evidence. This allows the system to distinguish between a genuine annotation error and a simple model failure, addressing the spatial reasoning gap explored in our research questions.

\begin{figure}[t]
\centering
\begin{minipage}[b]{1\textwidth}
    \centering
    \includegraphics[width=\textwidth, trim=10 0 80 0, clip=true]{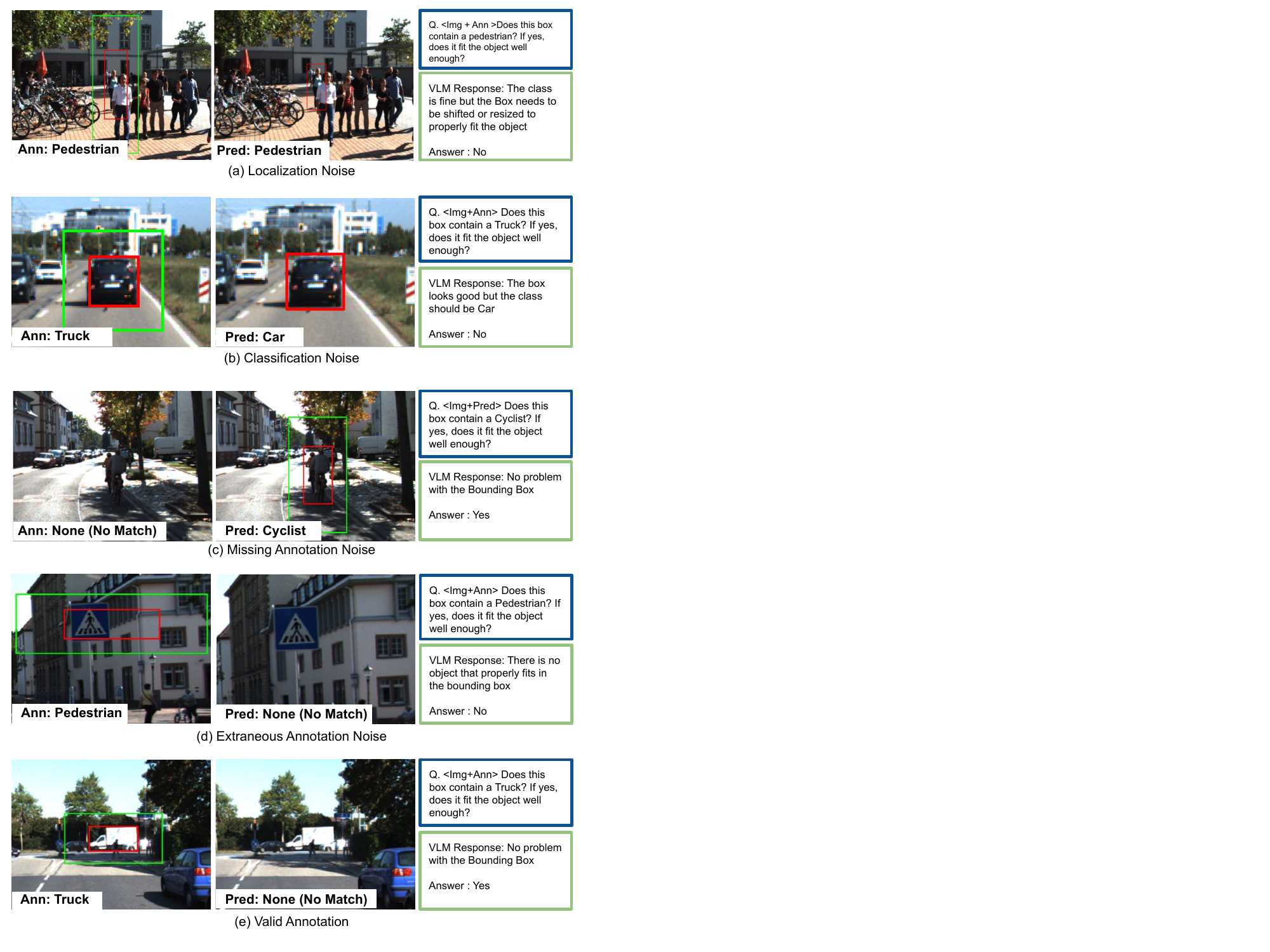}
\end{minipage}
\vspace{-0.7cm}
\caption{The examples demonstrate different noise types and the performance of fine-tuned with CoT VLM Llama on each noise type and valid annotations. The columns represent annotations, predictions, and Q\&A respectively.
The green box is the visual prompt $P^v$. 
}
\vspace{-0.5cm}
\label{fig:qual_combined}
\end{figure}

\section{Experiments} \label{sec:experiments}

We outline the experimental set up designed to answer the posed research questions for our study. We quantify performance throughout the study using standard metrics: Accuracy (Acc.), Recall (Rec.), Precision (Prec.), and the F1 Score (F1). Our study utilizes recall to measure error discovery and precision to ensure data integrity is maintained.

\subsection{Datasets and Noise Profiles}\label{sec:dataset}
We evaluate the framework on two standard 2D object detection benchmarks: KITTI ($D_{\mathrm{K}}$) \cite{Geiger2012CVPR} and nuImages ($D_{\mathrm{nu}}$) \cite{nuscenes}. Since these datasets are relatively clean and, we assume, lack substantial erroneous annotations, we introduce new known errors by randomly modifying the original annotations. To measure the error detection rate accurately, Fig.~\ref{fig:qual_combined}, we inject four noise types simulating common dataset decays: localization shifts, classification swaps, missing annotations, and extraneous boxes.

\textbf{KITTI}: This dataset consists of 63,160 bounding box annotations across nine classes, which we divided into 80:10:10 for training, validation, and testing subsets ($D_{\mathrm{K:train}}$, $D_{\mathrm{K:val}}$, $D_{\mathrm{K:test}}$).
We filter out non-semantic labels, like $\mathit{DontCare}$ and $\mathit{Misc}$, and also heavily occluded or truncated objects to ensure visual clarity for the VLM, 
The noise-induced version, $D_{\mathrm{K:test}}^{\mathrm{noise}}$, contains around 30\% of evenly distributed erroneous annotations.

\textbf{nuImages}: For larger-scale evaluation, we used nuImages, which contains over 800,000 bounding boxes and 23 foreground classes. We followed the official splits for training, validation, and testing ($D_{\mathrm{nu:train}}$, $D_{\mathrm{nu:val}}$, $D_{\mathrm{nu:test}}$). No filtering was applied to test AutoVDC on a more complex and diverse set of real-world data. 
The noise-injected versions, $D^{\mathrm{noise}}_{\mathrm{nu:train}}$ and $D^{\mathrm{noise}}_{\mathrm{nu:test}}$, contain around 15\% of evenly distributed injected noise.
\subsection{EP Stage: High-Recall Filtering}\label{sec:ep_stage_eval}
The Error Proposal (EP) stage utilizes DETR \cite{carion2020endtoendobjectdetectiontransformers} detector to generate candidates.

\textbf{KITTI Setup}: The task model trained on  clean set, $D_{\mathrm{K:train}}$, to establish a performance ceiling using a strong proposer.

\textbf{nuImages Setup}: Intentionally trained on a corrupted set ($D^{\mathrm{noise}}_{\mathrm{nu:train}}$) to mimic a realistic weak proposer where existing labels are already imperfect.

\begin{table}[t]
\caption{(\ref{sec:exp1}) Evaluation metrics of our system using different VLMs on $D_{\mathrm{K:test}}^{\mathrm{noise}}$. We see that a poor performer (Llama) can be drastically improved with fine-tuning (FT, FT-CoT).}
\label{table:kitti_eval}
\resizebox{\columnwidth}{!}{%
    \renewcommand{\arraystretch}{1.2} 
    \begin{tabular}{l|l|c|c|c|c}
    \Xhline{2\arrayrulewidth}
        Model & Eval. Type & Rec. & Prec. & Acc. & F1 \\
        \hline
        - & Error Proposals & \textbf{0.96} & 0.72 & 0.89 & 0.82 \\
        \hline
        \multirow{2}{*}{DeepSeek-VL2 \cite{deepseekvl2} } & Error Validation & 0.58 & \textbf{0.99} & 0.69 & 0.73 \\
         & AutoVDC System & 0.55 & \textbf{0.99} & 0.87 & 0.71 \\
        \hline
        \multirow{2}{*}{Gemini Flash 2.0 \cite{google_geminiflash2.0}} & Error Validation & 0.84 & 0.83 & 0.76 & 0.83 \\
         & AutoVDC System & 0.80 & 0.83 & 0.90 & 0.81 \\
        \hline
        \multirow{2}{*}{GPT 4.1 \cite{openai_chatgpt4.1}} & Error Validation & 0.84 & 0.84 & 0.77 & 0.84 \\
         & AutoVDC System & 0.80 & 0.84 & 0.90 & 0.82 \\
        \hline
        \multirow{2}{*}{ViP-LLaVA \cite{cai2024vipllavamakinglargemultimodal} } & Error Validation & 0.61 & 0.71 & 0.53 & 0.66 \\
         & AutoVDC System & 0.58 & 0.71 & 0.82 & 0.64 \\
        \hline
        \multirow{2}{*}{Llama \cite{grattafiori2024llama3herdmodels} } & Error Validation & 0.47 & 0.89 & 0.58 & 0.62 \\
         & AutoVDC System & 0.45 & 0.89 & 0.83 & 0.60 \\
        \hline
        \multirow{2}{*}{Llama (FT)} & Error Validation & 0.91 & 0.91 & 0.87 & 0.91 \\
         & AutoVDC System & 0.88 & 0.91 & 0.94 & 0.89 \\
        \hline
        \multirow{2}{*}{Llama (FT-CoT)} & Error Validation & \textbf{0.96} & 0.94 & \textbf{0.92} & \textbf{0.95} \\
         & AutoVDC System & \textbf{0.92} & 0.94 & \textbf{0.96} & \textbf{0.93} \\
         \Xhline{2\arrayrulewidth}
    \end{tabular}
}
\vspace{-0.4cm}
\end{table}

\subsection{EV Stage: Fine-Tuning and CoT parameters}\label{sec:finetuning}

We utilize the Llama-3.2-Vision-11B model  \cite{grattafiori2024llama3herdmodels} in both zero-shot mode, fine-tuned(FT) and fine-tuned with Chain of Thought (FT-CoT) modes. 
We evaluate a diverse selection of VLMs to provide a representative snapshot of the multi-modal reasoning maturity. 
To achieve dataset alignment:

\textbf{Fine-Tuning Split}: We re-purposed the validation sets from both of our datasets, $D_{\mathrm{K:val}}$ and $D_{\mathrm{nu:val}}$, by injecting 50\% noise to create noise-induced subsets, $D^{\mathrm{noise}}_{\mathrm{K:FT}}$ and $D^{\mathrm{noise}}_{\mathrm{nu:FT}}$ for fine-tuning.

\textbf{Training Setup}:  We employed LoRA~\cite{hu2022lora} for fine-tuning, updating the query and value projection modules. 
The input consisted of image with either noise-injected or clean annotations, and the target $Y$ indicated whether they are erroneous. 
In the FT-CoT variant, the target output $Y$ was augmented with a text-based justification.
 
For the text prompt, we use \textit{\enquote{Does this box contain the $\{\mathrm{class\_label}\}$. If yes, does it fit the object well enough?}}. 


\subsection{Evaluation Criteria}\label{sec:evaluation_criteria}


The \textbf{EP evaluation} addresses the triage logic of the framework. By comparing the error proposals $D^{\prime}$ against the ground truth ${D}$, we quantify the Error Proposal stage's ability to act as a high-recall filter.

The \textbf{EV evaluation} focuses on the reasoning capabilities of the VLM. By comparing the validated output $D^{\prime\prime}$ against the candidates provided by the first stage $D^{\prime}$, we can isolate the VLM's expert judgment to measure its ability to resolve the spatial reasoning gap and the effectiveness of domain-alignment, such as FT-CoT.

The \textbf{AutoVDC System Evaluation} assesses the end-to-end effectiveness. We compare the final validated errors $D^{\prime\prime}$ against the original dataset ${D}$ to evaluate performance
across varying noise scales.

The \textbf{Task Model Evaluation} demonstrates the practical impact on task development. By re-evaluating the perception model across clean ${D}$, noisy $D^{\mathrm{noise}}$, and cleaned $D^{\mathrm{cleaned}}$ datasets, we show how using VLMs can have a real impact on data curation and, in turn, restores evaluation fidelity.




\begin{table}[t]
\caption{(\ref{sec:exp4}) Evaluation on $D^{\mathrm{noise}}_{\mathrm{nu:test}}$ of AutoVDC with DETR trained with $D^{\mathrm{noise}}_{\mathrm{nu:train}}$ and Llama. Even with a weak task model, AutoVDC can still reliably find erroneous annotations. 
}
\label{table:nuimages_eval}
\resizebox{\columnwidth}{!}{%
  \renewcommand{\arraystretch}{1.2} 
    \begin{tabular}{l|l|c|c|c|c}
    \Xhline{2\arrayrulewidth}
        Model & Eval. Type & Rec. & Prec. & Acc. & F1 \\
        \hline
        - & Error Proposals & {0.86} & 0.19 & 0.43 & 0.31 \\
        \hline
        \multirow{2}{*}{Llama \cite{grattafiori2024llama3herdmodels} } & Error Validation & 0.51 & 0.30 & 0.68 & 0.38 \\
         & AutoVDC System & 0.44 & 0.30 & 0.76 & 0.36 \\
        \hline
        \multirow{2}{*}{Llama (FT-CoT)} & Error Validation & {0.89} & {0.78} & {0.93} & {0.83} \\
         & AutoVDC System & \textbf{0.77} & \textbf{0.78} & \textbf{0.93} & \textbf{0.77} \\
         \Xhline{2\arrayrulewidth}
    \end{tabular}
}
\vspace{-0.4cm}
\end{table}

\subsection{Experimental Results}\label{sec:evaluation_metrics}

In order to quantify the effectiveness of the proposed framework, we conduct the following experiments. 

\subsubsection{\textbf{VLM Sensitivity and the Spatial Reasoning Gap}}\label{sec:exp1}

Our first study investigates whether general-purpose foundation models can identify annotation errors out-of-the-box or if they require task-specific alignment. We use the DETR task model trained on $D_{\mathrm{K:train}}$ as our task model for the EP stage for this part of the study.

In Tab.~\ref{table:noise-type-eval}, zero-shot models such as Gemini Flash 2.0 and GPT-4.1 demonstrate decent semantic capabilities, achieving over 0.95 recall on classification noise. However, they exhibit a significant spatial reasoning gap.
In Tab.~\ref{table:kitti_eval}, DeepSeek-VL2 has high precision of 0.99, it has low recall of 0.55, indicating that it is unable to detect almost half the noise in the dataset. Tab.~\ref{table:noise-type-eval} clarifies this inability by showing that DeepSeek-VL2 is an expert at identifying missing annotations (0.99 recall) but it fails almost entirely on localization noise (0.01 recall).


As shown in Tab.~\ref{table:kitti_eval}, FT-CoT transforms the base Llama model from a low zero-shot performer (F1: 0.60) to our strongest validator (F1: 0.93). Most importantly, Tab.~\ref{table:noise-type-eval} highlights that FT-CoT recovers the missing spatial performance, raising localization recall from the baseline 0.14 to a robust 0.90. This confirms that for high-fidelity curation, domain-specific alignment through CoT reasoning is a fundamental requirement to resolve intricate spatial errors.

\subsubsection{\textbf{Resilience to Weak Candidate Proposals}}\label{sec:exp4}

We tested this by intentionally training a DETR model on a dataset with a high noise rate of 15\%, $D^{\mathrm{noise}}_{\mathrm{nu:train}}$. This created a weak proposer for the nuImages dataset. We evaluate using $D^{\mathrm{noise}}_{\mathrm{nu:test}}$.
As shown in Tab.~\ref{table:nuimages_eval}, the weak task model was very noisy, producing error proposals with a precision of only 0.19. 
However, 
our Llama (FT-CoT) was able to filter through this noise, maintaining a high validation precision of 0.78.
This proves the EP stage's role as Economic Triage
by flagging suspicious samples, it reduced the VLM's total workload substantially. This allowed the VLM to focus its expensive reasoning power only on the most likely errors. 

\begin{table}[t]
\caption{(\ref{sec:exp1}) Error Validation Recall metrics of different VLMs for each noise type on $D_{\mathrm{K:test}}^{\mathrm{noise}}$. The VLMs are particularly weak on handling localization errors, which benefits most from fine-tuning.
}
\label{table:noise-type-eval}
\resizebox{0.99\columnwidth}{!}{%
    \renewcommand{\arraystretch}{1.99} 
    \begin{tabular}{l|c|c|c|c}
        \Xhline{2\arrayrulewidth}
        \vspace{-0.05cm}
        Model & \makecell{Localization \\ Noise} & \makecell{Classification \\ Noise} & \makecell{Missing Ann. \\ Noise} & \makecell{Extraneous Ann.\\ Noise} \\
        \hline
        DeepSeek-VL2 \cite{deepseekvl2 } & 0.01 & 0.57 & \textbf{0.99} & 0.80 \\ 
        \hline
        Gemini Flash 2.0 \cite{google_geminiflash2.0} & 0.65 & 0.96  & 0.74 & 0.99 \\
        \hline
        GPT 4.1 \cite{openai_chatgpt4.1} & 0.59 & 0.95  & 0.80 & 1.00 \\
        \hline
        ViP-LLaVA \cite{cai2024vipllavamakinglargemultimodal} & 0.58 & 0.66  & 0.41 & 0.74 \\
        \hline
        Llama \cite{grattafiori2024llama3herdmodels} & 0.14 & 0.38 & 0.89 & 0.52 \\
        \hline
        Llama (FT) & 0.85 & 0.9 & 0.92 & 1.00 \\
        \hline
        Llama (FT-CoT) & \textbf{0.90} & \textbf{0.97} & {0.95} & \textbf{1.00} \\
        \Xhline{2\arrayrulewidth}
    \end{tabular}
}
\vspace{-0.4cm}
\end{table}

\subsubsection{\textbf{Real-World Discovery and Noise Scaling}}\label{sec:exp2}

\begin{figure}
    \centering    \includegraphics[width=1\linewidth]{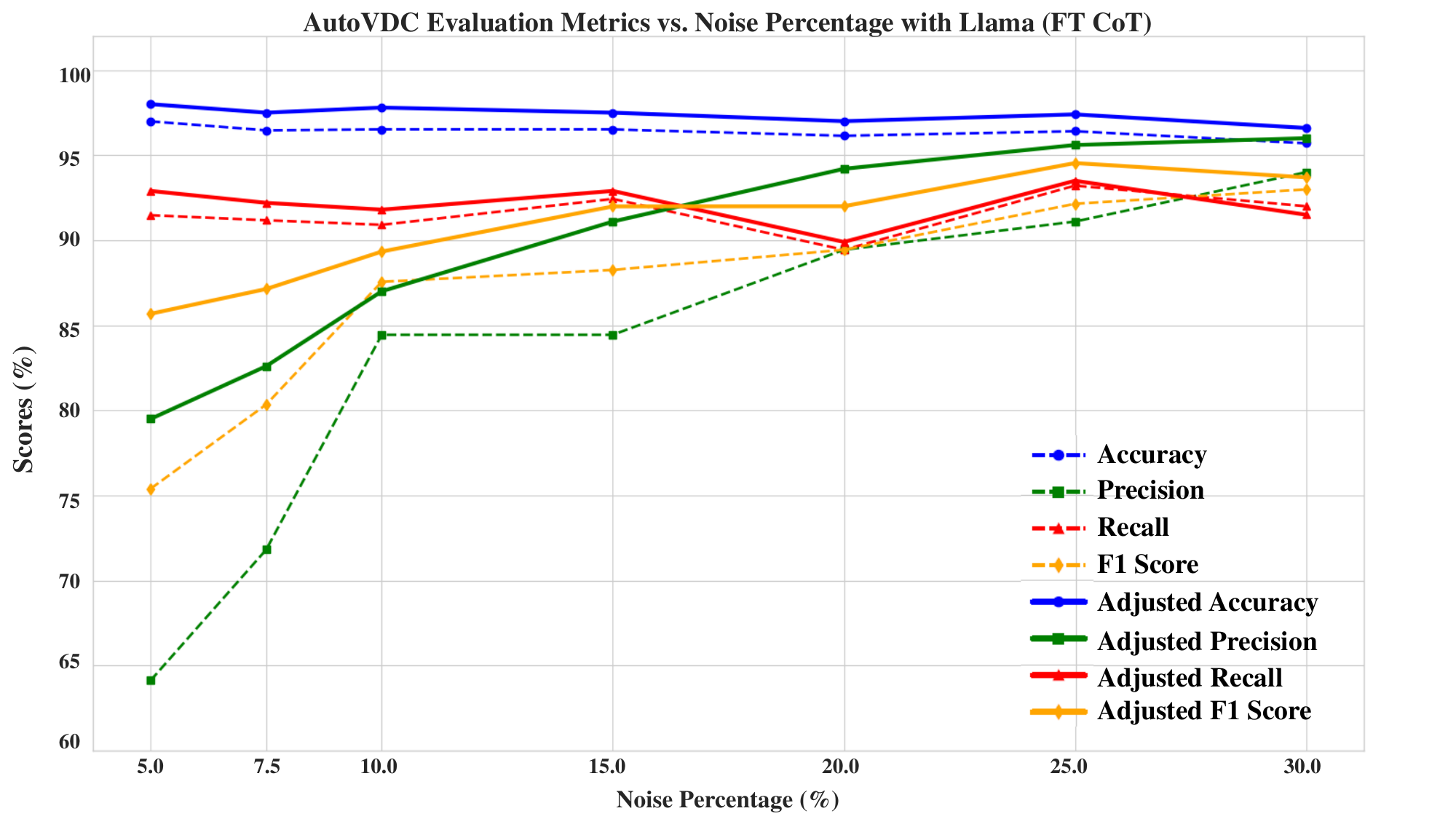}
    \caption{(\ref{sec:exp2}) Real, non-injected erroneous annotations can lead to biased results at lower noise rates, so removing the identified 39 real errors is critical for accurate assessment (Adjusted), but is mitigated by using larger noise rates.}
    \label{fig:noise-range}
\vspace{-0.7cm}
\end{figure}

We tested the performance on the KITTI test set with injected noise rates ranging from 5\% to 30\%, in $D_{\mathrm{K:test}}^{\mathrm{noise}}$. Llama (FT-CoT) model, Fig. \ref{fig:noise-range}, maintains consistent accuracy and recall metrics across the ranges. While precision naturally decreases at lower noise rates as there are fewer errors to find giving more weightage to the false positives, the overall performance remains stable.


Beyond synthetic noise, we also evaluated the system's ability to find actual errors in the original, unaltered KITTI dataset, $D_{\mathrm{K:test}}$. Even at a 0\% injected noise rate, AutoVDC identified 71 suspicious annotations, 39 of which were confirmed as genuine mistakes in the original ground truth, mainly included completely occluded objects, even though the dataset was filtered based on the occlusion annotation attribute. This proves that the study isn't just on artificial errors we added but it actually finds real mistakes that humans missed in prominent datasets.



For datasets with varying noise rates, Fig. \ref{fig:noise-range}, we report the performance metrics both before and after mitigating the effect of 39 erroneous annotations (\enquote{Adjusted}). 
Accounting for these 39 real errors significantly improves our performance metrics across all noise rates. 
Overall, we see that at all noise rates, AutoVDC keeps consistent accuracy and recall metrics while dropping precision at lower noise rates.

\subsubsection{\textbf{Impact on Task Model Evaluation}}\label{sec:exp3}



In Tab.~\ref{table:test_set_eval}, we established a baseline using the clean KITTI test set $D_{\mathrm{K:test}}$, where a DETR achieves 0.38 precision and 0.54 recall. The model's perceived performance collapses to 0.22 precision and 0.36 recall on $D_{\mathrm{K:test}}^{\mathrm{noise}}$. 
We created clean versions of the noisy dataset using three VLM variants: zero-shot Llama, fine-tuned Llama (FT), and fine-tuned Llama with Chain-of-Thought (FT-CoT).

$D_{\mathrm{K:test}}^{\mathrm{cleaned_{FT-CoT}}}$ had 696 erroneous annotations removed and regained 11\% to reach a precision of 0.33 and match the baseline's recall at 0.54.

\begin{table}[t!]
\caption{(\ref{sec:exp3}) Evaluation of DETR on cleaned up datasets along with a breakdown of which annotations were removed (\enquote{Ann. Rm.}) after cleaning KITTI test dataset with 758 noise-injected erroneous annotations. AutoVDC gets close to the \enquote{true} evaluation numbers.}
\label{table:test_set_eval}
\resizebox{0.99\columnwidth}{!}
{
\renewcommand{\arraystretch}{1.3} 
    \begin{tabular}{l|c|c|c|c}
        \Xhline{2\arrayrulewidth}
        Dataset & \multicolumn{2}{c|}{DETR Evaluation} & \# Erroneous & \# Non-Erroneous \\
        \cline{2-3}
        \vspace{-0.07cm}
         & Avg. Rec. & Avg. Prec. & Ann. Rm. & Ann. Rm.\\
        \hline
        ${D_{\mathrm{K:test}}}$ & 0.54 & 0.38 & - & - \\
        \hline
        ${D_{\mathrm{K:test}}^{\mathrm{noise}}}$ & 0.36 & 0.22 & - & - \\
        \hline
        ${D_{\mathrm{K:test}}^{\mathrm{cleaned}}}$ & {0.4} & {0.24} & {344} & 41 \\
        \hline
        ${D_{\mathrm{K:test}}^{\mathrm{cleaned_{FT}}}}$ & {0.52} & {0.32} & {666} & 67 \\
        \hline
        ${D_{\mathrm{K:test}}^{\mathrm{cleaned_{FT-CoT}}}}$ & \textbf{0.54} & {0.33} & {696} & 46 \\
        \Xhline{2\arrayrulewidth}
    \end{tabular}
}
\vspace{-0.5cm}
\end{table}

\section{Discussion}\label{sec:Discussion}


\textbf{VLM Readiness and the Spatial Gap}: Our findings in \ref{sec:exp1} provide a readiness report for current VLMs in automated data curation. We observe a clear hierarchy in task difficulty. 
While zero-shot VLMs are highly mature for semantic classification, a significant spatial reasoning gap persists. 
As shown in Tab.~\ref{table:nuimages_eval}, foundation models cannot yet reliably audit fine-grained localization without specific alignment. 
The need for extensive fine-tuning may decrease as the native spatial reasoning of industry VLMs improves.



\textbf{The Efficiency of Triage Logic}: A core contribution of the AutoVDC framework is the two-stage methodology that exploits the generalization gap of neural networks. As demonstrated in \ref{sec:exp4}, the system is resilient to a weak proposer. Even with an EP Stage precision of only 0.19, the VLM acts as a robust gatekeeper, maintaining high system precision. This economic triage logic is vital for scalability, as it reduces the human auditing workload exponentially and making high-fidelity curation feasible for massive production datasets.

\textbf{Human-in-the-Loop and the Long-Tail}: As seen in Tab.~\ref{table:test_set_eval}, while FT-CoT is the most precise validator, removing only 46 valid samples compared to the 67 removed by standard fine-tuning, we recognize that automated cleaning cannot yet totally replace human oversight. These removed samples often represent the long-tail of the dataset, cases that are difficult or ambiguous that they appear erroneous even to advanced models. 
The discovery of 39 genuine, non-synthetic errors in the original KITTI ground truth empirically validates that the FT-CoT reasoning generalizes beyond injected patterns to capture real-world dataset decay. 
Rather than eliminating manual review, AutoVDC transforms it from an exhaustive search into a targeted verification of high-confidence candidates.

\textbf{Systematic Bias}:
A limitation of the current approach is systemic label bias. 
 If a dataset contains consistent, repetitive error patterns, the task model may learn them as features, preventing the EP stage from flagging them as discrepancies.
Addressing systematic label biases represents a promising area for future research.

\textbf{Noise Modeling}: 
We view noise modeling as an adjustable parameter. 
This helps the finetuned VLM understand the specific characteristics of the dataset during testing.
We can then better align the VLM's error detection capabilities with the nuances of the dataset, thereby improving its performance.


\textbf{Methodological Extension}: 
To evaluate cross-task modularity, we conceptually extended AutoVDC to 3D Bird's Eye View (BEV) occupancy task. By projecting 3D voxel discrepancies into the 2D perspective domain, we enabled the VLM to act as a cross-view auditor, successfully identifying ghost objects and missed obstacles in a preliminary proof-of-concept.
This provides a robust road-map for future research, suggesting that general-purpose VLMs can act as universal auditors for complex perception tasks, provided the errors are correctly mapped into the model's native visual space.


\section{Conclusion}\label{sec:conclusion}

This work introduced AutoVDC, a modular framework designed to evaluate the maturity of Vision-Language Models (VLMs) for automated data curation. By employing a two-stage architecture for our study, leveraging the latent generalization of task models to propose error candidates and the reasoning power of VLMs for final validation, we successfully bridge the gap between model intuition and ground-truth integrity.


Our systematic study reveals that while zero-shot VLMs are highly effective for semantic classification, specialized Chain-of-Thought (CoT) fine-tuning is essential for resolving fine-grained localization noise. 
Furthermore, we proved that the generalization gap of task models provides a reliable signal for noise discovery even when trained on corrupted labels. 
Our experiments confirm that the framework remains robust when guided by weak task models, with the VLM acting as a high-precision gatekeeper that reduces manual review workloads exponentially.

Instead of auditing the entire dataset or even the full list of error proposals, manual verifiers can focus exclusively on the high-confidence candidates flagged by the VLM. This significantly cuts down on human effort and expense while ensuring that the long-tail samples are preserved. Ultimately, this methodology offers a robust roadmap for shifting toward automated, high-fidelity data engines across diverse vision-based applications.
{
  \small
  \bibliographystyle{IEEEtran}
  \bibliography{Literature}

@IEEEtranBSTCTL{IEEEexample:BSTcontrol,
CTLuse_article_number = "yes",
CTLuse_paper = "yes",
CTLuse_forced_etal = "yes",
CTLmax_names_forced_etal = "6",
CTLnames_show_etal = "3",
CTLuse_alt_spacing = "yes",
CTLalt_stretch_factor = "4",
CTLdash_repeated_names = "yes",
CTLname_latex_cmd = "",
CTLname_url_prefix = ""
}

@inproceedings{vqa2015,
  title={{VQA}: {V}isual {Q}uestion {A}nswering},
  author={Antol, Stanislaw and Agrawal, Aishwarya and Lu, Jiasen and Mitchell, Margaret and Batra, Dhruv and Zitnick, C Lawrence and Parikh, Devi},
  booktitle={ICCV},
  year={2015}
}

@inproceedings{nuscenes,
  title={nu{S}cenes: A multimodal dataset for autonomous driving},
  author={Caesar, Holger and Bankiti, Varun and Lang, Alex H and Vora, Sourabh and Liong, Venice Erin and Xu, Qiang and Krishnan, Anush and Pan, Yu and Baldan, Giancarlo and Beijbom, Oscar},
  booktitle={CVPR},
  year={2020}
}

@inproceedings{lewis1994sequential,
title={A sequential algorithm for training text classifiers},
author={Lewis, David D and Gale, William A},
booktitle={SIGIR},
year={1994},
organization={Springer London}
}

@techreport{settles2009active,
title={Active learning literature survey},
author={Settles, Burr},
year={2009},
institution={University of Wisconsin-Madison Department of Computer Sciences},
number={TR 1648}
}

@inproceedings{lakshminarayanan2017simple,
title={Simple and Scalable Predictive Uncertainty Estimation using Deep Ensembles},
author={Lakshminarayanan, Balaji and Pritzel, Alexander and Blundell, Charles},
booktitle={NeurIPS},
year={2017}
}

@book{bishop2006pattern,
title={Pattern Recognition and Machine Learning},
author={Bishop, Christopher M},
year={2006},
publisher={Springer}
}

@inproceedings{ramaswamy2000efficient,
title={Efficient algorithms for mining outliers from large data sets},
author={Ramaswamy, Sridhar and Rastogi, Rajeev and Shim, Kyuseok},
booktitle={SIGMOD/PODS},
year={2000}
}

@misc{radford2021learningtransferablevisualmodels,
title={Learning Transferable Visual Models From Natural Language Supervision}, 
author={Alec Radford and Jong Wook Kim and Chris Hallacy and Aditya Ramesh and Gabriel Goh and Sandhini Agarwal and Girish Sastry and Amanda Askell and Pamela Mishkin and Jack Clark and Gretchen Krueger and Ilya Sutskever},
year={2021},
eprint={2103.00020},
archivePrefix={arXiv},
}

@misc{xiao2023florence2advancingunifiedrepresentation,
title={Florence-2: Advancing a Unified Representation for a Variety of Vision Tasks}, 
author={Bin Xiao and Haiping Wu and Weijian Xu and Xiyang Dai and Houdong Hu and Yumao Lu and Michael Zeng and Ce Liu and Lu Yuan},
year={2023},
eprint={2311.06242},
archivePrefix={arXiv},
}

@misc{ravi2024sam2segmentimages,
title={{SAM} 2: Segment Anything in Images and Videos}, 
author={Nikhila Ravi and Valentin Gabeur and Yuan-Ting Hu and Ronghang Hu and Chaitanya Ryali and Tengyu Ma and Haitham Khedr and Roman Rädle and Chloe Rolland and Laura Gustafson and Eric Mintun and Junting Pan and Kalyan Vasudev Alwala and Nicolas Carion and Chao-Yuan Wu and Ross Girshick and Piotr Dollár and Christoph Feichtenhofer},
year={2024},
eprint={2408.00714},
archivePrefix={arXiv},
}

@misc{agrawal2024pixtral12b,
title={Pixtral 12B}, 
author={Pravesh Agrawal and Szymon Antoniak and Emma Bou Hanna and Baptiste Bout and Devendra Chaplot and Jessica Chudnovsky and Diogo Costa and Baudouin De Monicault and Saurabh Garg and Theophile Gervet and Soham Ghosh and Amélie Héliou and Paul Jacob and Albert Q. Jiang and Kartik Khandelwal and Timothée Lacroix and Guillaume Lample and Diego Las Casas and Thibaut Lavril and Teven Le Scao and Andy Lo and William Marshall and Louis Martin and Arthur Mensch and Pavankumar Muddireddy and Valera Nemychnikova and Marie Pellat and Patrick Von Platen and Nikhil Raghuraman and Baptiste Rozière and Alexandre Sablayrolles and Lucile Saulnier and Romain Sauvestre and Wendy Shang and Roman Soletskyi and Lawrence Stewart and Pierre Stock and Joachim Studnia and Sandeep Subramanian and Sagar Vaze and Thomas Wang and Sophia Yang},
year={2024},
eprint={2410.07073},
archivePrefix={arXiv},
}

@misc{liu2023visualinstructiontuning,
      title={Visual Instruction Tuning}, 
      author={Haotian Liu and Chunyuan Li and Qingyang Wu and Yong Jae Lee},
      year={2023},
      eprint={2304.08485},
      archivePrefix={arXiv},
      primaryClass={cs.CV},
}

@misc{abdin2024phi3technicalreporthighly,
      title={Phi-3 Technical Report: A Highly Capable Language Model Locally on Your Phone}, 
      author={Marah Abdin and Jyoti Aneja and Hany Awadalla and Ahmed Awadallah and Ammar Ahmad Awan and Nguyen Bach and Amit Bahree and Arash Bakhtiari and Jianmin Bao and Harkirat Behl and Alon Benhaim and Misha Bilenko and Johan Bjorck and Sébastien Bubeck and Martin Cai and Qin Cai and Vishrav Chaudhary and Dong Chen and Dongdong Chen and Weizhu Chen and Yen-Chun Chen and Yi-Ling Chen and Hao Cheng and Parul Chopra and Xiyang Dai and Matthew Dixon and Ronen Eldan and Victor Fragoso and Jianfeng Gao and Mei Gao and Min Gao and Amit Garg and Allie Del Giorno and Abhishek Goswami and Suriya Gunasekar and Emman Haider and Junheng Hao and Russell J. Hewett and Wenxiang Hu and Jamie Huynh and Dan Iter and Sam Ade Jacobs and Mojan Javaheripi and Xin Jin and Nikos Karampatziakis and Piero Kauffmann and Mahoud Khademi and Dongwoo Kim and Young Jin Kim and Lev Kurilenko and James R. Lee and Yin Tat Lee and Yuanzhi Li and Yunsheng Li and Chen Liang and Lars Liden and Xihui Lin and Zeqi Lin and Ce Liu and Liyuan Liu and Mengchen Liu and Weishung Liu and Xiaodong Liu and Chong Luo and Piyush Madan and Ali Mahmoudzadeh and David Majercak and Matt Mazzola and Caio César Teodoro Mendes and Arindam Mitra and Hardik Modi and Anh Nguyen and Brandon Norick and Barun Patra and Daniel Perez-Becker and Thomas Portet and Reid Pryzant and Heyang Qin and Marko Radmilac and Liliang Ren and Gustavo de Rosa and Corby Rosset and Sambudha Roy and Olatunji Ruwase and Olli Saarikivi and Amin Saied and Adil Salim and Michael Santacroce and Shital Shah and Ning Shang and Hiteshi Sharma and Yelong Shen and Swadheen Shukla and Xia Song and Masahiro Tanaka and Andrea Tupini and Praneetha Vaddamanu and Chunyu Wang and Guanhua Wang and Lijuan Wang and Shuohang Wang and Xin Wang and Yu Wang and Rachel Ward and Wen Wen and Philipp Witte and Haiping Wu and Xiaoxia Wu and Michael Wyatt and Bin Xiao and Can Xu and Jiahang Xu and Weijian Xu and Jilong Xue and Sonali Yadav and Fan Yang and Jianwei Yang and Yifan Yang and Ziyi Yang and Donghan Yu and Lu Yuan and Chenruidong Zhang and Cyril Zhang and Jianwen Zhang and Li Lyna Zhang and Yi Zhang and Yue Zhang and Yunan Zhang and Xiren Zhou},
      year={2024},
      eprint={2404.14219},
      archivePrefix={arXiv},
      primaryClass={cs.CL},
}

@article{Bernhardt_2022,
   title={Active label cleaning for improved dataset quality under resource constraints},
   volume={13},
   ISSN={2041-1723},
   DOI={10.1038/s41467-022-28818-3},
   number={1},
   journal={Nature Communications},
   publisher={Springer Science and Business Media LLC},
   author={Bernhardt, Mélanie and Castro, Daniel C. and Tanno, Ryutaro and Schwaighofer, Anton and Tezcan, Kerem C. and Monteiro, Miguel and Bannur, Shruthi and Lungren, Matthew P. and Nori, Aditya and Glocker, Ben and Alvarez-Valle, Javier and Oktay, Ozan},
   year={2022},
}

@misc{li2024llmsasjudgescomprehensivesurveyllmbased,
      title={LLMs-as-Judges: A Comprehensive Survey on LLM-based Evaluation Methods}, 
      author={Haitao Li and Qian Dong and Junjie Chen and Huixue Su and Yujia Zhou and Qingyao Ai and Ziyi Ye and Yiqun Liu},
      year={2024},
      eprint={2412.05579},
      archivePrefix={arXiv},
      primaryClass={cs.CL},
}

@misc{guo2024generativeaisyntheticdata,
      title={Generative {AI} for Synthetic Data Generation: Methods, Challenges and the Future}, 
      author={Xu Guo and Yiqiang Chen},
      year={2024},
      eprint={2403.04190},
      archivePrefix={arXiv},
      primaryClass={cs.LG},
}

@article{Chen_2022,
   title={Automatic Labeling to Generate Training Data for Online LiDAR-Based Moving Object Segmentation},
   volume={7},
   number={3},
   journal={RA-L},
   author={Chen, Xieyuanli and Mersch, Benedikt and Nunes, Lucas and Marcuzzi, Rodrigo and Vizzo, Ignacio and Behley, Jens and Stachniss, Cyrill},
   year={2022},
}

@misc{zheng2023judgingllmasajudgemtbenchchatbot,
      title={Judging LLM-as-a-Judge with MT-Bench and Chatbot Arena}, 
      author={Lianmin Zheng and Wei-Lin Chiang and Ying Sheng and Siyuan Zhuang and Zhanghao Wu and Yonghao Zhuang and Zi Lin and Zhuohan Li and Dacheng Li and Eric P. Xing and Hao Zhang and Joseph E. Gonzalez and Ion Stoica},
      year={2023},
      eprint={2306.05685},
      archivePrefix={arXiv},
}

@misc{carion2020endtoendobjectdetectiontransformers,
      title={End-to-End Object Detection with Transformers}, 
      author={Nicolas Carion and Francisco Massa and Gabriel Synnaeve and Nicolas Usunier and Alexander Kirillov and Sergey Zagoruyko},
      year={2020},
      eprint={2005.12872},
      archivePrefix={arXiv},
}

@inproceedings{Geiger2012CVPR,
  author = {Andreas Geiger and Philip Lenz and Raquel Urtasun},
  title = {Are we ready for Autonomous Driving? The {KITTI} Vision Benchmark Suite},
  booktitle = {CVPR},
  year = {2012}
}

@inproceedings{hu2022lora,
title={Lo{RA}: Low-Rank Adaptation of Large Language Models},
author={Edward J Hu and Yelong Shen and Phillip Wallis and Zeyuan Allen-Zhu and Yuanzhi Li and Shean Wang and Lu Wang and Weizhu Chen},
booktitle={ICLR},
year={2022},
}

@misc{cai2024vipllavamakinglargemultimodal,
      title={{ViP-LLaVA}: Making Large Multimodal Models Understand Arbitrary Visual Prompts}, 
      author={Mu Cai and Haotian Liu and Dennis Park and Siva Karthik Mustikovela and Gregory P. Meyer and Yuning Chai and Yong Jae Lee},
      year={2024},
      eprint={2312.00784},
      archivePrefix={arXiv},
      primaryClass={cs.CV},
}

@InProceedings{Sun_2020_CVPR, author = {Sun, Pei and Kretzschmar, Henrik and Dotiwalla, Xerxes and Chouard, Aurelien and Patnaik, Vijaysai and Tsui, Paul and Guo, James and Zhou, Yin and Chai, Yuning and Caine, Benjamin and Vasudevan, Vijay and Han, Wei and Ngiam, Jiquan and Zhao, Hang and Timofeev, Aleksei and Ettinger, Scott and Krivokon, Maxim and Gao, Amy and Joshi, Aditya and Zhang, Yu and Shlens, Jonathon and Chen, Zhifeng and Anguelov, Dragomir}, title = {Scalability in Perception for Autonomous Driving: Waymo Open Dataset}, booktitle = {CVPR}, month = {June}, year = {2020} }

@inproceedings{wilson2023argoverse2generationdatasets,
  author = {Benjamin Wilson and William Qi and Tanmay Agarwal and John Lambert and Jagjeet Singh and Siddhesh Khandelwal and Bowen Pan and Ratnesh Kumar and Andrew Hartnett and Jhony Kaesemodel Pontes and Deva Ramanan and Peter Carr and James Hays},
  title = {Argoverse 2: Next Generation Datasets for Self-driving Perception and Forecasting},
  booktitle = {NeurIPS Datasets and Benchmarks},
  year = {2021}
}

@misc{schuhmann2022laion5bopenlargescaledataset,
      title={LAION-5B: An open large-scale dataset for training next generation image-text models}, 
      author={Christoph Schuhmann and Romain Beaumont and Richard Vencu and Cade Gordon and Ross Wightman and Mehdi Cherti and Theo Coombes and Aarush Katta and Clayton Mullis and Mitchell Wortsman and Patrick Schramowski and Srivatsa Kundurthy and Katherine Crowson and Ludwig Schmidt and Robert Kaczmarczyk and Jenia Jitsev},
      year={2022},
      eprint={2210.08402},
      archivePrefix={arXiv},
      primaryClass={cs.CV},
}

@misc{heyn2023automotiveperceptionsoftwaredevelopment,
      title={Automotive Perception Software Development: An Empirical Investigation into Data, Annotation, and Ecosystem Challenges}, 
      author={Hans-Martin Heyn and Khan Mohammad Habibullah and Eric Knauss and Jennifer Horkoff and Markus Borg and Alessia Knauss and Polly Jing Li},
      year={2023},
      eprint={2303.05947},
      archivePrefix={arXiv},
      primaryClass={cs.SE}
}

@misc{liu2024surveyautonomousdrivingdatasets,
      title={A Survey on Autonomous Driving Datasets: Statistics, Annotation Quality, and a Future Outlook}, 
      author={Mingyu Liu and Ekim Yurtsever and Jonathan Fossaert and Xingcheng Zhou and Walter Zimmer and Yuning Cui and Bare Luka Zagar and Alois C. Knoll},
      year={2024},
      eprint={2401.01454},
      archivePrefix={arXiv},
      primaryClass={cs.CV}
}

@misc{grattafiori2024llama3herdmodels,
      title={The Llama 3 Herd of Models}, 
      author={Aaron Grattafiori et al.},
      year={2024},
      eprint={2407.21783},
      archivePrefix={arXiv},
      primaryClass={cs.AI},
}

@misc{google_geminiflash2.0,
  author = {Google},
  title = {Gemini Flash 2.0},
  note = {Accessed on April 2025},
}

@misc{openai_chatgpt4.1,
  author = {OpenAI},
  title = {ChatGPT (GPT-4.1)},
  note = {Accessed on April 2025},
}

@misc{deepseekvl2,
      title={DeepSeek-VL2: Mixture-of-Experts Vision-Language Models for Advanced Multimodal Understanding}, 
      author={Zhiyu Wu and Xiaokang Chen and Zizheng Pan and Xingchao Liu and Wen Liu and Damai Dai and Huazuo Gao and Yiyang Ma and Chengyue Wu and Bingxuan Wang and Zhenda Xie and Yu Wu and Kai Hu and Jiawei Wang and Yaofeng Sun and Yukun Li and Yishi Piao and Kang Guan and Aixin Liu and Xin Xie and Yuxiang You and Kai Dong and Xingkai Yu and Haowei Zhang and Liang Zhao and Yisong Wang and Chong Ruan},
      year={2024},
      eprint={2412.10302},
      archivePrefix={arXiv},
      primaryClass={cs.CV},
}

@misc{lu2025clipgraderleveragingvisionlanguagemodels,
      title={Clip{G}rader: Leveraging Vision-Language Models for Robust Label Quality Assessment in Object Detection}, 
      author={Hong Lu and Yali Bian and Rahul C. Shah},
      year={2025},
      eprint={2503.02897},
      archivePrefix={arXiv},
      primaryClass={cs.CV},
}

@InProceedings{Kim_2024_CVPR,
    author    = {Kim, Junsu and Ku, Yunhoe and Kim, Jihyeon and Cha, Junuk and Baek, Seungryul},
    title     = {VLM-PL: Advanced Pseudo Labeling Approach for Class Incremental Object Detection via Vision-Language Model},
    booktitle = {CVPR Workshops},
    year      = {2024},
}

@misc{diab2025vlmsforsemisupervisedlearning,
author = {Diab, Mohanad and Barchi, Grazia and Moser, David},
year = {2025},
month = {01},
title = {Vision-Language Models as Pseudo-Label Validators in Semi-Supervised Learning: Geo-Locating Medium Voltage Cabins from Google Street View Imagery},
doi = {10.2139/ssrn.5167011}
}
}
\clearpage 
\onecolumn 

\section*{Appendix}
\section{Erroneous Annotations in the Original KITTI Dataset}
\label{sec:appendix_kitti_errors}

\begin{figure}[h!]
    \centering
    \includegraphics[width=0.65\textwidth , trim=10 65 370 0, clip=true]{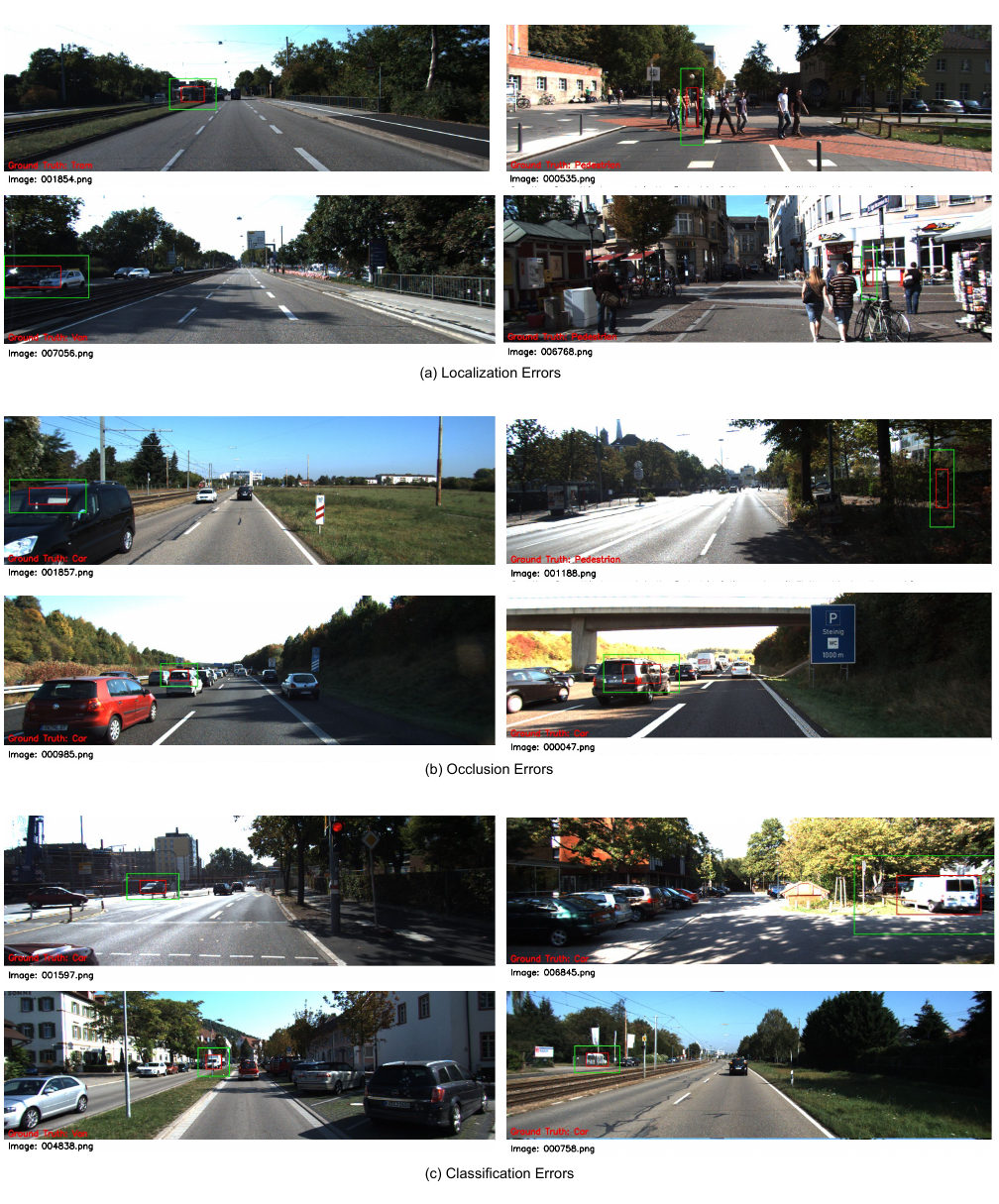}
    \caption{Examples of genuine annotation errors found in the original KITTI test set, as discussed in Appendix~\ref{sec:appendix_kitti_errors}. The figure displays 12 samples categorized into three common error types: inaccurate bounding box placement (Localization), annotations for fully occluded objects (Occlusion), and incorrect class assignments (Classification). These errors were identified by our AutoVDC system.}
    \label{fig:kitti_errors}
\end{figure}

During our analysis, even after filtering, we identified 39 annotations in the original KITTI test set that were genuinely erroneous. These pre-existing errors were consistently flagged by our AutoVDC framework across different noise rates, as discussed in Section \ref{sec:exp2}. To provide insight into the nature of these errors, Figure \ref{fig:kitti_errors} showcases 12 representative examples from this set of 39 incorrect annotations.

The figure is organized by error type to highlight common failure modes in manual or semi-automated annotation pipelines. We feature four examples for each of the following categories:
\begin{itemize}
    \item \textbf{Localization Errors:} The bounding boxes are inaccurately placed, either being poorly centered, incorrectly sized, or shifted, failing to tightly enclose the object of interest.
    \item \textbf{Occlusion-based Errors:} These annotations correspond to objects that are almost completely occluded and not visible in the camera image, making them invalid for vision-based detection tasks despite potentially being present in the 3D scene data.
    \item \textbf{Classification Errors:} The class label assigned to the bounding box is incorrect (e.g., a 'Van' labeled as a 'Car').
\end{itemize}

These examples underscore the necessity of automated data cleaning tools like AutoVDC, as even widely-used benchmark datasets are not entirely free of annotation noise.

\end{document}